\documentclass[a4paper,12pt]{article}
\usepackage{graphicx}
\begin{document}
\title{A Definition of Artificial Intelligence}
\author{Dimiter Dobrev\\
Institute of Mathematics and Informatics\\
Bulgarian Academy of Sciences\\
Sofia 1090, BULGARIA\\
e-mail: d@dobrev.com}
\renewcommand{\today}{January 19, 2004}
\maketitle

\begin{abstract}
In this paper we offer a formal definition of Artificial Intelligence and this directly gives us an algorithm for construction of this object. Really, this algorithm is useless due to the combinatory explosion. 

The main innovation in our definition is that it does not include the knowledge as a part of the intelligence. So according to our definition a newly born baby also is an Intellect. Here we differs with Turing's definition which suggests that an Intellect is a person with knowledge gained through the years.
\end{abstract}
\section*{1. Introduction}

This paper is about one basic problem. This is the problem for defining the notion of Artificial Intelligence. It is surprising that such basic problem can be still open. For example from a long time we have a definition of the notion of the computer. As such definition can be accepted the Turing's machine [6]. Not long time after the definition of the computer the first computer was made.

The same person Alan Turing made the most widely spread definition of AI. This is the so called Turing's test [7, 8, 9]. It is quite simple. We place something behind a curtain and it speaks with us. If we can't make difference between it and a human being then it will be AI. However, this definition is not formal. Another problem is that this definition does not separate the knowledge from the intellect. Imagine that you give a definition of the computer which does not separate the software from the hardware. Such definition would sound something like: ``Computer is a box and when you switch on the power you see windows and buttons. It include some nice games. You can also use it to watch movies.'' Such definition of a computer defines something much more complicated than the Turing's machine. It is much easier to built a computer following the Turing's definition than by following the second one.

\section*{2. Definition of AI}

We will offer a new formal definition of AI. In this definition we are going to exclude the knowledge from the intelligence and define something that knows nothing but which can learn. So according to our definition a newly born baby is also an Intellect.

Before giving a formal definition of AI we will make three acceptable assumptions. First assumption is the thesis of Church [1], stating that every calculating device can be modelled by a program. This means that we are going to look for AI in the set of programs. Second assumption is that AI is a step device\footnote{illustrations - Konstantin Lakov} and on every step it inputs from outside a portion of information (a letter from a finite alphabet $\Sigma$) and outputs a portion of information (a letter from a finite alphabet $\Omega$). The third assumption is that AI is in some environment which gives it a portion of information on every step and which receives the output of AI. Also we assume that the environment will be influenced of the information which AI outputs. This environment can be natural or artificial and we will refer to it as ``world''.

\begin{figure}

\begin{center}
\includegraphics[width=50mm]{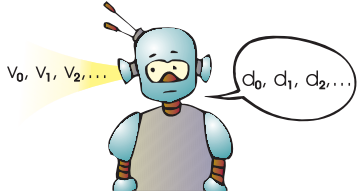}
\end{center}
  \caption{The step device}
\end{figure}

Now we can state informally our definition: {\bf AI will be such a program which in an arbitrary world will cope not worse than a human.}

In order to formalise this definition we need to formalise the notion of ``world'' and to say when one program copes in one world better than another. First, what is a world for us? These will be one set $S$, one element $s_{0}$ of $S$ and two functions $World(s, d)$ and $View(s)$. The set $S$ contains the internal states of the world and it can be finite or infinite. The element $s_{0}$ of $S$ will be the world's starting state. The function $World$ will take as arguments the current state of the world and the influence that our device has on the world at the current step. As a result, this function will return the new state of the world (which it will obtain on the next step). The function $View$ will inform us what does our device see. An argument of this function will be the world's state and the returned value will be the information that the device will receive (at a given step). We can suppose that the function $View$ is inaction but this assumption is too strong because in this case the set $S$ has to be finite and because in this case AI see all in its world. For example for us this is not true. We do not see behind our back.

If we have a world and a program then we can start it in this world. We will say that the program is living in this world. The life will start from the state $s_{0}$. This will be the world's state when our program was born. During its life the world will go through the states $s_{0},\ s_{1},\ s_{2},\ \ldots\ $. The program will influence the world with the information it works out at each step $d_{0},\ d_{1},\ d_{2},\ \ldots\ $. Also, our program will receive information from the world $v_{0},\ v_{1},\ v_{2},\ \ldots\ $. It is clear that $s_{i+1}=World(s_{i} , d_{i})$ and $v_{i}=View(s_{i})$.

To obtain a better idea for the world let us define the tree of the obtainable states. This will be infinite tree with countably many knots where every knot has $k$ inheritors. Here $k$ is the number of all possible actions (the number of  the letters in $\Omega$). To the tree's root we are going to juxtapose the state $s_{0}$. This world's state will be reached at the moment of birth. To the inheritors of the root we will juxtapose the states $World(s_{0}, d_{i})$ where $d_{i}$ runs through the alphabet $\Omega$. These states can be reached in a moment one (if the action in moment zero was the respective one). By analogy, we continue with the inheritors of the inheritors and so on. In this way to every knot of the tree we juxtapose one obtainable state of the world. Of course, one state can be juxtaposed to more than one knot.

\begin{figure}
\begin{center}
\includegraphics[width=50mm]{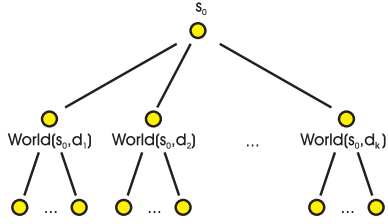}
\end{center}
  \caption{The tree of the obtainable states}
\end{figure}

On figure 3 you can see a more rough picture of this tree. In this figure only two knots are denoted. This is the moment of birth and the present moment. The path between these two knots we will call ``the life'' or ``life experience until the present moment''.

\begin{figure}
\begin{center}
\includegraphics[width=50mm]{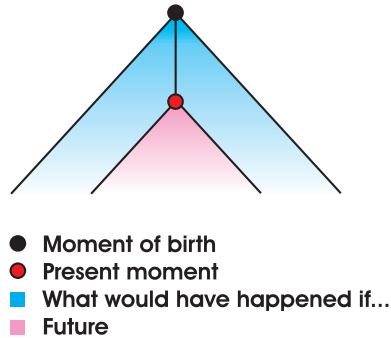}
\end{center}
  \caption{Rough picture of the tree of the obtainable states}
\end{figure}

From the tree of the obtainable states we can easily get another tree, which we will call ``the tree of the life''. This will be the same tree but at each knot instead some state $s_{i}$ we will juxtapose $View(s_{i})$, i.g. instead the respective world's state we will juxtapose the information the device gets as an entrance when it is in that state (what it sees). Why did we call this tree with the pretentious name ``tree of the life''? It is because it describes completely the current life of the device together with all possible variant for the past and for the future. If we have two different worlds and if they have the same tree of the life then these worlds are absolutely indistinguishable from the point of view of the device. No matter what experiment it would carry out, it would get the same result in both worlds because with the same sequence of actions it would see the same things.

One interesting question is whether we consider the function $World$ as determined or not. The answer is that it does not matter because we consider that we live our life only once and we cannot check on the second time is it determined or not. It would be better to ask is there a dependence which determines the function $World$. If we do not know this dependence then we can consider that such dependence does not exist and that the function $World$ is random. For example, the semi-random numbers generated by the computer are not random but they are generated by enough difficult dependence so we can consider them random. Also if we have real random numbers then we can consider that they are generated by some very complicated dependence which we do not know.

Our next goal will be to compare two lives and say which one is better. This means to define a linear order in the set of finite rows $v_{0},\ v_{1},\ \ldots\ ,\ v_{t}$. We will compare only finite rows because every life is finite and even if it is potentially unfinite then we will compare it until the present moment because we do not know what will happen in the future. Our order will not depend from the rows $d_{0},\ d_{1},\ \ldots\ ,\ d_{t}$ and $s_{0},\ s_{1},\ \ldots\ ,\ s_{t}$ because it does not matter what we do and what are the actual states of the world in which we are. The only thing that matters is what we view as a result of our activity.

We will choose one linear order of the set of finite rows $v_{0},\ v_{1},\ \ldots\ ,\ v_{t}$ and we will call this order the meaning of life. We suppose that the meaning of life is given beforehand. The reason for that is that we cannot hope that our device will cope well in one world without knowing what is to cope well. So the meaning of life is given beforehand and we do not expect that AI will find it itself. With the humans the situation is similar. They receive the meaning of life by some instincts and by the education.

To simplify the definition we will choose one concrete meaning of life. We will suppose that alphabet $\Sigma$ has two prior given subsets $\Sigma_{1}$ and $\Sigma_{2}$. Let $\Sigma_{1}$ will be the subset of the good things and $\Sigma_{2}$ will be the subset of the bad things. We will evaluate one life $v_{0},\ v_{1},\ \ldots\ ,\ v_{t}$ with the number of the good things in it minus the number of the bad thing in it. We will say that one life is better than another if its value is bigger, i.e. if in this life we saw more good thing and less bad. We do not suggest that the intersection of $\Sigma_{1}$ and $\Sigma_{2}$ is empty but if one element is in both $\Sigma_{1}$ and $\Sigma_{2}$ then it is the same as if it was not in any of them.

Now our definition is almost formal because we formalized the world and the meaning of the life. The only thing which is not formal is that we compare AI with a human being. We cannot say simply that in any world AI copes well because there are worlds in which no one can cope well. Imagine that the function $World(s, d)$ do not depend on $d$. In this case it will happen the same dose not matter what we do. In such a world everybody will cope equally. Also we have to suppose that there are not fatal errors in the world. This mean that we give enough time for education to our device. Other problem is that the world can be too complicated. Of course, we can suppose that AI is more intelligent than any human being and that if one human will manage in one world then AI will manage too. Anyway, for any program we can find world which is enough complicated so the program cannot cope in it.

We will say that one world is good if there are not fatal errors in it and if it is not too complicated for a human being.

\section*{3. Algorithm for searching of AI}

If we had a formal definition of AI we would have an algorithm for searching of AI. The reason is that the set of programs is countable and if we have decidable or semi-decidable test which to recognize AI then we can start checking all programs one by one until we find AI. (In the case of semi-decidable test the algorithm is a little bit more complicated.) Really, such algorithm is useless due to the combinatory explosion but anyway, the existence of such algorithm is interesting.

Although our definition is not completely formal we can make a test for intelligence and this means that we can make an algorithm for searching of AI. The idea is the same as the student exams. We give them several tasks and consider intelligent this students who manage with all tasks. In our test the tasks will be good worlds which are artificial (programs made by people). (Two examples of artificial world can be found in the examples of the compiler Strawberry Prolog [2].) We will start the candidate program in such world and give it enough time for education. After that time we will see how well it copes in this world and does it cover requirements for this world (for example, in the next hundred steps the relation victory to loss to be at least 9 to 1). 

If the requirements are not too tough for the human and if AI exists then it will pass our test. The problem is that AI is not the only program which will pass this test. For any finite number of good worlds there is a program which copes in these worlds but which is not AI. For example, if this program is written especially for the test worlds. We have the same problem with the students' exams. Many people who have learned all the tasks by heart will pass the exam but this people are not intellects but crammers.

Actually, what we propose is not a test for AI but if worlds included in this test are enough numerous and varied, then the shortest program which will pass it will be AI. (Because the crammer program will be more complicated). We will consider that our algorithm orders the programs according to their length. So the first (the simplest) which will be worked out from our algorithm will be AI.

As we said the algorithm described above for searching for AI is entirely useless due to the combinatory explosion but it is not so with the definition of AI. After learning what is AI we can try to build it directly. Really, even if we have AI then we cannot use it directly because first we have to train it. The same is with the computer. The hardware is nothing without software. Anyway the training will be not a problem because we have big experience with training people.


\section*{References}


.

[1] C\,h\,u\,r\,c\,h, A. (1941) {\it The Calculi of Lambda-Conversion.} Princeton: Princeton University Press

[2] D\,o\,b\,r\,e\,v D. {\it Strawberry Prolog}, http://www.dobrev.com

[3] D\,o\,b\,r\,e\,v D. {\it AI Project}, http://www.dobrev.com/AI

[4] D\,o\,b\,r\,e\,v D. (2000) {\it AI - What is this}, PC Magazine - Bulgaria, November'2000

[5] D\,o\,b\,r\,e\,v D. (2001) {\it AI - How does it cope in an arbitrary world}, PC Magazine - Bulgaria, February'2001

[6] T\,u\,r\,i\,n\,g, A. M. (1936) {\it On Computable Numbers, with an Application to the Entscheidungsproblem.} Proceedings of the London Mathematical Society, Series 2, 42 (1936-37), pp.230-265.

[7] T\,u\,r\,i\,n\,g, A. M. (1948) {\it Intelligent machinery}, report for National Physical Laboratory, in Machine Intelligence 7, eds. B. Meltzer and D. Michie (1969) 

[8] T\,u\,r\,i\,n\,g, A. M. (1950) {\it Computing machinery and intelligence}, Mind 49: pp 433-460

[9] T\,u\,r\,i\,n\,g, A. M. (1956) {\it Can a Machine Think}, in volume 4 of The World of Mathematics, ed. James R. Newman, pp 2099-2123, Simon \& Schuster


\end{document}